\documentclass[a4paper,twoside]{article}

\usepackage{cite}
\usepackage{microtype}
\usepackage{amsmath,amssymb,amsfonts}
\usepackage{algorithmic}
\usepackage{graphicx}
\usepackage{textcomp}
\usepackage{xcolor}
\usepackage{url}
\usepackage{placeins}
\usepackage{tabularx}
\usepackage[hidelinks]{hyperref}
\usepackage{dblfloatfix}
\usepackage{booktabs}
\usepackage{threeparttable}
\usepackage{booktabs, 
            multirow, threeparttable}  
\usepackage{siunitx}                    
\usepackage{xparse} 
\usepackage[table]{xcolor}
\usepackage{epsfig}
\usepackage{subcaption}
\usepackage{calc}
\usepackage{amssymb}
\usepackage{amstext}
\usepackage{amsmath}
\usepackage{amsthm}
\usepackage{multicol}
\usepackage{pslatex}
\usepackage{apalike}
\usepackage{algorithm2e}
\usepackage[bottom]{footmisc}
\usepackage{SCITEPRESS}     
\usepackage{cite}
\usepackage{amsmath,amssymb,amsfonts}
\usepackage{algorithmic}
\usepackage{graphicx}
\usepackage{textcomp}
\usepackage{xcolor}
\usepackage{booktabs}
\usepackage{makecell}

\definecolor{LimeBlue}{RGB}{95,155,175}
\definecolor{KshapOrange}{RGB}{245,165,95}
\definecolor{FaGreen}{RGB}{110,170,105}

\newcommand{\LIME}[2]{\cellcolor{LimeBlue!#1}#2}
\newcommand{\KSHAP}[2]{\cellcolor{KshapOrange!#1}#2}
\newcommand{\FA}[2]{\cellcolor{FaGreen!#1}#2}

\begin{document}

\title{Evaluating Local Explainability Metrics for Machine Learning Models on Tabular Data}

\author{\authorname{Tom\'as Pereira\sup{1}\orcidAuthor{0009-0004-7764-6691}, Jo\~ao Vitorino\sup{1}\orcidAuthor{0000-0002-4968-3653}, Eva Maia \sup{1}\orcidAuthor{0000-0002-8075-531X} and Isabel Pra\c{c}a\sup{1}\orcidAuthor{0000-0002-2519-9859}}
\affiliation{\sup{1}GECAD, ISEP, Polytechnic of Porto, Porto, Portugal}
}


\abstract{Despite the wide use of explainability techniques to attempt to understand the behavior of Artificial Intelligence (AI), the generated explanations may not always be reliable. An explanation can appear plausible to humans but fail to capture the internal reasoning of a model, particularly when dealing with complex tabular data.
This paper studies the trustworthiness of local explainability techniques when applied to complex tabular classification tasks, considering evaluated metrics for three main properties: faithfulness to the model’s predictions, robustness to input data variations, and complexity of the explanation itself.
A benchmark was performed for Local Interpretable Model-Agnostic Explanations (LIME), Kernel SHapley Additive exPlanations (SHAP), and Feature Ablation techniques, across 32 datasets and different types of machine learning models.
Model performance ranges were analyzed to identify two groups: consensus-correct, which are samples that all models predicted correctly, and consensus-wrong, samples that all models predicted incorrectly. The obtained results demonstrate that that the explanations are not always correlated with a model's predictive performance. Instead, dataset complexity and feature distributions seem to be the main factors affecting explanation quality and reliability.}

\onecolumn \maketitle \normalsize \setcounter{footnote}{0} \vfill

\section{Introduction}
Artificial Intelligence (AI) is being increasingly used to support and automate decisions and business processes across a wide range of domains. In particular, machine learning models are now frequently used to automate tasks such as risk assessment, fraud detection, medical decision support, and other operational processes where predictions directly influence real-world outcomes \cite{SAEED2023110273,adadi}. However, with the increasing complexity of AI models, in most cases users cannot trace how a predictions was made. These black-box models prioritize performance and accuracy over explainability and unlike more simple models like linear models, their inner workings are a lot of times unknown \cite{SAEED2023110273,adadi,stop}.

To address this challenge, a broad range of explainability techniques have been proposed. These techniques can have a global or a local scope \cite{Zhang_2021}. Global methods attempt to explain the model's behavior as a whole as means to verify if their decision boundaries are adequate to the problem at hand\cite{Zhang_2021}. On the other hand, local explanations are particularly relevant in risk-sensitive scenarios, where understanding a specific decision is essential for trust, debugging, and accountability \cite{accountability}.

Despite the growing number of explainability techniques, evaluating the quality of an explanation itself remains an open problem. An explanation can appear plausible to humans but fail to represent the model's true behavior. Furthermore, the same method may not always be consistent, behaving differently depending on the predictive model, dataset characteristics, or input data. Therefore it is important to perform a systematic and reproducible evaluation of the usefulness of the commonly used explainability techniques, with quantitative evaluation metrics.

This work studies the trustworthiness of local explainability techniques when applied to complex tabular classification tasks, aiming to identify patterns and correlations between dataset complexity, model performance, and explanation trustworthiness. A benchmark was performed for three commonly used techniques: Local Interpretable Model-Agnostic Explanations (LIME), Kernel SHapley Additive exPlanations (SHAP), and Feature Ablation. The considered evaluation metrics were focused on three main properties: faithfulness to the model's predictions, robustness to input data variations, and complexity of the explanation itself.

The remainder of this paper is organized as follows. Section \ref{related_work} reviews related work and compares available frameworks. Section \ref{methods} describes the selected metrics, explainability techniques, models, datasets, and the evaluation process. Section \ref{results} presents and discusses the results. Finally, Section \ref{conclusions} summarizes the main conclusions and future work.

\section{Related Work} \label{related_work}

It is essential to review prior work in explainable artificial intelligence (XAI), identifying which explainability techniques have been most widely adopted in recent years, and which evaluation metrics are more suitable for complex tabular data.

\subsection{Explainability Techniques}

A review of recent papers and surveys addressing explanation techniques in XAI was performed. with the goal of identifying which have been most frequently adopted in recent years and which of them are repeatedly recommended for practical machine-learning settings. Across the surveyed literature, three broad categories are defined \cite{Zhang_2021}, \cite{coutner}: 
\begin{itemize}
    \item Global: Global explanations try to explain the general behavior of the model. They are useful to, for example, determining the feature importance of the models.
    \item Local: This type of explanation aims to explain a single prediction. They are critical to understand why the model takes certain decisions. 
    \item Counterfactual: Counterfactual explanations are mainly local explanations that try to understand what perturbations are necessary in the features of a single instance for the model's prediction to change.
\end{itemize}

Many different explainability techniques exist, covering various model architectures and goals. Table \ref{table:Explanations} summarizes the main explanation families considered in this study.

\begin{table}[h]
  \centering
  \resizebox{\columnwidth}{!}{%
    \begin{threeparttable}
      \caption{Overview of Explainability Techniques}
      \label{table:Explanations}
      \begin{tabular}{p{2.7cm}|c|c}
        \hline
        Technique  & Type & Model \\ \hline
        LIME \cite{lime} & Local & Agnostic\\
        SHAP \cite{shap} & Local/Global\tnote{1} & Machine Learning\\
        TreeSHAP \cite{treeSHAP} & Local/Global\tnote{1} & Tree models\\
        LinearSHAP \cite{shap} & Local/Global\tnote{1} & Linear models\\
        GradientSHAP \cite{shap} & Local/Global\tnote{1} & Neural Networks\\
        Kernel SHAP \cite{shap} & Local/Global\tnote{1} & Agnostic\\
        Feature Ablation \cite{fa} & Local/Global\tnote{1} & Agnostic\\
        Integrated Gradients \cite{ig} & Local & Neural Networks \\
        Grad-CAM \cite{gradcam} & Local & Neural Networks\\
        Partial Dependency Plots \cite{pdp} & Global & Machine Learning\\
        DiCE \cite{dice} & Counterfactual/Local & Agnostic\\
        \hline
      \end{tabular}
      \begin{tablenotes}
        \item[1] Global with sample aggregation 
      \end{tablenotes}
    \end{threeparttable}%
  }
\end{table}

LIME is a local, model-agnostic technique that explains a single prediction by generating perturbed samples around the instance of interest and fitting an interpretable surrogate model, typically linear, in that local neighborhood. SHAP is a framework based on Shapley values from cooperative game theory, where each feature receives a score according to its contribution to the prediction. Depending on how results are aggregated, SHAP can be interpreted locally for a single sample or globally across a dataset using sample aggregation.

Several other specialized techniques exist to efficiently adapt SHAP to various models. TreeSHAP adapts SHAP to tree-based models and LinearSHAP performs the same role for linear models, making use of their simpler structure. GradientSHAP extends the idea to neural networks by combining gradient-based information with SHAP attribution values.

Kernel SHAP is the model-agnostic implementation of SHAP and is especially relevant for this work because it can be applied to many model families. Feature Ablation is another model-agnostic approach that explains predictions by systematically removing or replacing individual features and measuring how much the model output changes. In practice, this provides a direct estimate of each feature's influence on the prediction.

The remaining techniques in the table cover settings that are outside the main focus of this paper but help give context to the chosen explanations. Integrated Gradients attributes predictions by accumulating gradients from a baseline input to the target instance, which makes it particularly suitable for neural network models. Grad-CAM highlights salient spatial regions in convolutional neural networks and is mainly used in image analysis. Partial Dependence Plots provide global explanations by showing how changes in one or more features affect the model's average prediction. Finally, DiCE generates counterfactual explanations by identifying minimal feature changes that would alter the predicted class.

In this paper, the focus is placed on local explanations, as they provide insight into model decisions at the sample level. Even though this paper deals only with machine learning models, only model agnostic explanations were chosen as it could later be expanded to use neural networks.

From these, LIME, Kernel SHAP and Feature Ablation were chosen as they are widely used and fit all criteria.

From the surveyed literature, several libraries exist in order to produce these explanations. LIME and SHAP have their own libraries where the authors created the base implementation of these techniques \cite{lime,shap}. However, several other model interpretability libraries exist that aggregate many of these techniques in one place \cite{coroama2022evaluation}. Captum, TorchRay and AIX360 are explainability libraries repeatedly referenced in the surveyed literature \cite{coroama2022evaluation,kadir2023evaluation,hedstrom2023quantus}. Captum is the official PyTorch explainability framework and provides multiple techniques, from LIME and SHAP to neural-network-oriented approaches such as Gradient SHAP and Integrated Gradients. This flexibility allows it to support multiple model and data types, including tabular and image data. TorchRay, in contrast, focuses on CNN explanation for image tasks through visualization-based methods.

\subsection{Explanation Evaluation Metrics}
Recent surveys and reviews on XAI evaluation show that explanation-assessment methods are typically organized into a small number of recurring metric families, including faithfulness, robustness, complexity, localization, axiomatic properties, and randomization-based tests \cite{coroama2022evaluation,kadir2023evaluation,hedstrom2023quantus}. These studies also emphasize that metric selection should depend on the data type, the explanation format, and the intended application, since many metrics were originally proposed for image-based or neural-network settings and are not always directly suitable for tabular machine-learning tasks \cite{coroama2022evaluation,kadir2023evaluation}. A comparison between the most relevant frameworks is presented in Table \ref{table:FrameworkComparison}.

\begin{table*}
  \centering
  \resizebox{\textwidth}{!}{%
  \begin{threeparttable}
    \caption{Comparison of XAI Evaluation Frameworks}
    \label{table:FrameworkComparison}
    \begin{tabular}{c c c c c c c c}
      \toprule
      Framework
      & \multicolumn{6}{c}{Property}
      & Supported Explanations \\
      \cmidrule(lr){2-7} \\ 
      & Faithfulness 
      & Robustness 
      & Localisation 
      & Complexity 
      & Axiomatic 
      & Randomisation  \\
      \midrule
      Quantus \cite{hedstrom2023quantus}  & 9 & 4 & 6 & 3 & 3 & 2 & Explanation-agnostic\tnote{a} \\
      AIX360 \cite{aix360-sept-2019}   & 2 & 0 & 0 & 0 & 0 & 0 & Explanation-agnostic\tnote{a} \\
      TorchRay \cite{torchray} & 0 & 0 & 1 & 0 & 0 & 0 & Explanation-agnostic\tnote{a} \\
      Captum \cite{kokhlikyan2020captum}   & 1 & 1 & 0 & 0 & 0 & 0 & Explanation-agnostic\tnote{a}\\
      LEAF \cite{leaf}     & 2 & 1 & 1 & 0 & 0 & 0 & LIME and SHAP \\
      \bottomrule
    \end{tabular}
    \begin{tablenotes}
      \item[a] Works for any generated explanation represented as a feature-importance tensor
    \end{tablenotes}
  \end{threeparttable}
  }
\end{table*}

As highlighted in recent surveys, most available XAI libraries and frameworks, such as Captum, TorchRay and AIX360, focus primarily on generating explanations rather than rigorously evaluating explanation quality \cite{coroama2022evaluation,kadir2023evaluation}. LEAF is a smaller library that does focus on explanation evaluation, however, it's limited to LIME and SHAP.

Among the surveyed libraries, Quantus provides the broadest range of evaluation metrics and is explanation-agnostic, as long as the generated explanations are represented as feature-importance vectors/tensors. In addition, Quantus supports Captum-based explanations natively, which simplifies integration in the evaluation pipeline. This choice is also supported by previous work using Quantus in image-based settings. In particular, \cite{bommer2023tutorial} presents a practical tutorial on Quantus using climate-related image data and neural networks, showing that the framework has already been applied successfully to the evaluation of explanations in that domain. Since several Quantus metrics have therefore already been explored and validated in image-oriented contexts, it becomes equally relevant to investigate how applicable those same metrics are to complex tabular data.

One practical challenge, also emphasized in the review literature, is that many existing metrics were originally designed for image settings \cite{coroama2022evaluation,kadir2023evaluation}. A general overview of explanation properties is shown in Table \ref{table:MetricGroups}.

\begin{table}[h]

\caption{Overview of Explanation Properties}
\label{table:MetricGroups}
\centering
\resizebox{\linewidth}{!}{%
\begin{tabular}{c|p{6.4cm}}
\hline
Property  & \multicolumn{1}{c}{Short Description} \\ \hline
Faithfulness  & How much the feature importance of the explanation correlates                                      with the actual model behavior.            \\
Robustness    & How much do perturbations in the inputs influence the explanation. \\
Localization  & Evaluates if the evidence of the explanation is centered around a certain region of interest (RoI)            \\
Complexity    & Evaluates how many features are used to explain the model prediction            \\
Axiomatic     & Assesses if explanations fulfill certain axiomatic properties            \\
Randomization & Evaluates how much randomizing the models parameters influence the explanation                     results            \\ \hline
\end{tabular}
}
\end{table}

As shown, different properties assess different aspects of explanation quality. Faithfulness focuses on whether the features identified as important by an explanation are truly aligned with model behavior. In recent surveys and reviews, this family is commonly associated with metrics such as Faithfulness Estimate, Faithfulness Correlation, Monotonicity, and Selectivity \cite{coroama2022evaluation,kadir2023evaluation,hedstrom2023quantus}. For complex tabular data, faithfulness metrics are particularly relevant because feature perturbations can be performed directly at the attribute level, making it possible to verify whether changes in the explanation are consistent with changes in model predictions.

Robustness evaluates whether small perturbations in the input produce stable explanations. In the literature, this property is frequently measured with Sensitivity-based metrics, including Average Sensitivity and Max Sensitivity, as well as related stability-oriented measures such as the Local Lipschitz Estimate \cite{coroama2022evaluation,kadir2023evaluation,hedstrom2023quantus}. These metrics also make sense for complex tabular data because they help determine whether explanations remain stable in the presence of small local changes.

Localization assesses whether the evidence highlighted by an explanation is concentrated in a relevant region of interest. Typical localization metrics are mostly based on overlap, ranking, or attention over spatial regions \cite{coroama2022evaluation,kadir2023evaluation}. Although they are useful in image tasks, they are generally less meaningful for tabular data, since tabular instances do not naturally contain spatial structure.

Complexity measures how concentrated or distributed an explanation is across the input features. This property often includes entropy-based and sparsity-oriented metrics \cite{coroama2022evaluation,kadir2023evaluation,hedstrom2023quantus}. For complex tabular data, these metrics are especially useful because they indicate whether a prediction is explained by a small number of relevant features or by a highly distributed attribution pattern, which directly affects interpretability.

Axiomatic properties assess whether explanations satisfy theoretical requirements such as completeness, sensitivity, or implementation invariance, depending on the explanation family under study \cite{coroama2022evaluation,kadir2023evaluation}. These metrics are important from a theoretical perspective, but they are often tied to specific classes of explanation techniques and are therefore less suitable for a broad comparison across different model-agnostic approaches.

Randomization evaluates whether explanations change when model parameters, labels, or internal structures are randomized, and is often used as a sanity-check in the literature \cite{coroama2022evaluation,kadir2023evaluation,hedstrom2023quantus}. However, these tests are usually more useful in deep learning scenarios, where parameter structure is central to the explanation process.

Therefore, to evaluate explanations in complex tabular classification tasks, the most relevant properties are faithfulness, robustness, and complexity. To the best of our knowledge, no previous work has performed a complete benchmark of these properties across a large quantity of tabular datasets.

\section{Methods} \label{methods}
This section details the evaluation process, including the considered metrics, datasets, explainability techniques, and machine learning models. It was carried out on a machine with a 14-core central processing unit and 16 gigabytes of random access memory.
\subsection{Evaluation Metrics}
In this work, the most relevant properties of the explanations were evaluated with a total of five different metrics:

\subsubsection{Faithfulness Estimate}
From the faithfulness metrics available in Quantus, two metrics were selected as being well suited for tabular classification. The first is Faithfulness Estimate \cite{faithest}, which assesses how well explanation attributions align with model behavior by perturbing features and measuring the corresponding change in prediction. Faithfulness varies between -1 and 1 with the highest value being complete alignment with the models decision boundary while -1 means all the model every feature the explanation considered important for the model's prediction was in reality not, and vice-versa.

\subsubsection{Selectivity}
The second is Selectivity \cite{select}, which evaluates whether features ranked as most important by the explanation actually drive the prediction when removed in order of importance. When applied to images, this technique is called Pixel Flipping. In Quantus, Selectivity is an image oriented metric. As such, in this implementation, each sample was converted into a 4D tensor so that it could be interpreted in Quantus and each feature could be used as if it were a pixel. In this way, the same effect can be accomplished by replacing the features with a baseline one by one. Selectivity values vary between 0 and 1, with a lower value signifying a higher alignment with the model.

\subsubsection{Average Sensitivity}
Average Sensitivity is derived from Sensitivity \cite{sens}, which measures how much an explanation changes under small input perturbations. A neighborhood radius is defined around each sample, and $n$ perturbed instances are generated through Monte Carlo sampling. Explanations for the perturbed samples are then compared against the original explanation. Average Sensitivity reports the mean change across perturbed samples and therefore captures the typical robustness of an explanation under small local variations. In this implementation, 20 perturbations were created with a lower bound of 0,01 and an upper bound of 0,05. Sensitivity values have a lower bound of 0 but do not have a defined upper bound. The lower the value, the more robust the explanation is.

\subsubsection{Max Sensitivity}
Max Sensitivity is based on the same perturbation process described above, but instead of measuring the average change, it reports the largest observed change across the perturbed samples. This metric is similar to Average Sensitivity, but it remains important to consider because it captures worst-case instability. Even when the average behavior appears acceptable, a single large deviation may indicate that an explanation is highly sensitive to specific perturbations, which is particularly relevant in high-stakes settings. As with Average Sensitivity, lower values indicate greater robustness.

\subsubsection{Complexity}
Complexity \cite{comp} measures how distributed an explanation is across features. It computes each feature's attribution share and then applies Shannon entropy. For example, if only two features have equal importance, each has a 50$\%$ share. Higher entropy corresponds to less sparse, more complex explanations. Complexity values vary between 0 and the natural logarithm of the number of features of each dataset. The lower the complexity value, the less complex the explanation is.

\subsection{Datasets and Data Preparation}

Due to the complex nature of tabular datasets, that can covering many different domains and data types, as well as all the different ways to treat said data, it is important to run a comprehensive benchmark in order to correctly evaluate the metrics in various different contexts.
TabArena\cite{tabarena} is a project that provides several high-quality, tabular machine learning datasets. Starting from over 1000 datasets, the authors filtered duplicates, non-tabular, synthetic datasets, and low-quality entries, yielding 51 curated datasets. In this study, we selected several classification datasets  spanning diverse domains, sample sizes, feature counts, and numbers of classes.

Table \ref{table:DatasetInfo} specifies the general characteristics of the used datasets, including minimum (Min.), maximum (Max.) and average (Avg.) values for the number of samples, features, numerical (Num.) and categorical (Cat.) features and missing values. In the appropriate fields, both the count (Nº) and the rate (in percentage) of that property in the dataset will be specified. Note that the dataset with the minimum/maximum number may not be the same as the one with the minimum/maximum rate.
\begin{table}[h]
\caption{Overview of dataset characteristics}
\label{table:DatasetInfo}
\centering
\begin{tabular}{c c c c}
\hline
Characteristic & Min. & Max. & Avg. \\ \hline
No. Samples & 748 & 76000 & 9777.871\\
No. Classes & 2 & 8 & 2.419 \\
No. Features & 4 & 1776 & 92.548 \\
Numerical (\%) & 0\% & 100\% & 62\% \\
Categorical (\%) & 0\% & 100\% & 37\% \\
Missing Values (\%) & 0\% & 8\% & 0.4\% \\

\bottomrule
\end{tabular}
\end{table}

All datasets were split into training and test sets using an 80\%/20\% ratio, with stratification by the target class to preserve class distributions. For preprocessing, numerical features were normalized, categorical features were one-hot encoded, and missing values were imputed using the median for numerical variables and the most frequent value for categorical variables.

\subsection{Models and Fine-tuning}
Three widely used model families with different complexity levels were selected to analyze whether explanation-metric behavior depends on predictor structure. Logistic Regression represents a simple linear baseline with high interpretability. Random Forest and XGBoost represent ensemble tree methods with stronger nonlinear modeling capacity.

Models were trained using Optuna \cite{optuna} for 30 optimization trials each and MLflow \cite{mlflow} was used for experiment tracking and result reproducibility. The utilized models and hyperparameters were the following:

\subsubsection{Logistic Regression}
Logistic Regression was selected as a simple and interpretable linear baseline. It provides a useful point of comparison against the tree-based models, making it possible to assess whether explanation metrics behave differently for models with relatively transparent decision boundaries. The search focused on the regularization strength while keeping the solver fixed to \texttt{lbfgs}, with \texttt{max\_iter=1000}. Table \ref{table:lr_grid} summarizes the explored search space.

\begin{table}[h]
\centering
\caption{Logistic Regression Hyperparameter Search Space}
\label{table:lr_grid}
\begin{tabular}{p{3.5cm}|p{3.2cm}}
\hline
Hyperparameter & Search Range / Values \\ \hline
Max Iterations & 1000 \\
Regularization strength $C$ & 0.1--10 \\
$l_1$ ratio & 0 \\
Solver & lbfgs \\ \hline
\end{tabular}
\end{table}

\subsubsection{Random Forest}
Random Forest was included as a robust ensemble baseline capable of modeling nonlinear relationships and feature interactions. Its aggregation of multiple decision trees also makes it less sensitive to variance than a single tree. The search focused on the number of trees and maximum depth, while also tuning split and leaf constraints to control overfitting. Table \ref{table:rf_grid} summarizes the explored search space.

\begin{table}[h]
\centering
\caption{Random Forest Hyperparameter Search Space}
\label{table:rf_grid}
\begin{tabular}{p{3.5cm}|p{3.2cm}}
\hline
Hyperparameter & Search Range / Values \\ \hline
Number of trees & 25--150 (step 25) \\
Maximum depth & None or 3--7 \\
Maximum features & sqrt, 5, or 10 \\
Min. samples split & 2--5 \\
Min. samples leaf & 1--2 \\ \hline
\end{tabular}
\end{table}

\subsubsection{XGBoost}
XGBoost was chosen as a stronger boosted-tree model with high predictive capacity, allowing the study to compare explanation behavior on a more powerful ensemble. The search emphasized parameters that most strongly affect boosting models, such as depth, learning rate, number of trees, and row/column subsampling. In addition, the objective was adapted to the number of target classes: \texttt{multi:softprob} with \texttt{mlogloss} for multiclass tasks, and \texttt{binary:logistic} with \texttt{logloss} for binary tasks. Table \ref{table:xgb_grid} summarizes the explored search space.

\begin{table}[h]
\centering
\caption{XGBoost Hyperparameter Search Space}
\label{table:xgb_grid}
\begin{tabular}{p{3.5cm}|p{3.2cm}}
\hline
Hyperparameter & Search Range / Values \\ \hline
Maximum depth & 2--4 \\
Learning rate & 0.03--0.10 \\
Number of trees & 50--300 \\
Subsample & 0.7--1.0 \\
Column sample by tree & 0.7--1.0 \\ \hline
\end{tabular}
\end{table}

\subsection{Explainability Techniques}
All explainability techniques were implemented with Captum and configured for local feature-attribution analysis.

\subsubsection{LIME}
LIME is a local, model-agnostic technique that explains a single prediction by fitting an interpretable surrogate model around the instance of interest. It generates perturbed samples in a local neighborhood, queries the original model on those samples, and then fits a weighted linear approximation. In this work, ridge regression is used as the surrogate model and 200 Gaussian-perturbed samples are generated per instance. Similarity between original and perturbed samples is computed using a Gaussian kernel with width 0.1 to preserve locality.

\subsubsection{Kernel SHAP}
Kernel SHAP (K-SHAP) is a model-agnostic approximation of Shapley values. Like LIME, it fits a local linear surrogate, but it uses SHAP-specific sampling and weighting to estimate each feature's marginal contribution. Instead of distance-based perturbations alone, SHAP relies on feature masking relative to a baseline to quantify attributions. In this work, 200 perturbed samples were generated to train the linear surrogate model.

\subsubsection{Feature Ablation}
Feature Ablation (FA) assumes that important features cause larger prediction changes when removed. It iteratively replaces features with a baseline and measures the change in model output. In this work, with normalized features, the per-feature mean is used as baseline.

\subsection{Evaluation Process}
The evaluation process was designed in a way that could relate explanation quality not only to the explanation technique itself, but also to model performance, predictive agreement, and dataset complexity. For each dataset, the three models were first trained and evaluated on the same holdout test split. Performance was measured with the F1-score, since this metric provides a balanced assessment of classification quality and remains informative even when class distributions are not perfectly uniform. Additionally, for each model, recall, precision and accuracy metrics were also stored.

In order to better understand the implications of the model's pedicion on explanation quality, two consensus groups were defined. The first, \textit{consensus-correct}, contains samples for which all models predicted the true class correctly. The second, \textit{consensus-wrong}, contains samples for which all models predicted incorrectly. These consensus groups were considered because they allow explanation behavior to be analyzed under two very different predictive conditions: one in which all models agree with the ground truth, and one in which all models fail simultaneously. This distinction is useful for determining whether explanation reliability is associated with correct model reasoning or whether explanations may still appear stable even when the underlying prediction is wrong.

Within each dataset, and for both groupings, five samples per target class were then selected whenever possible. This selection was performed only once, using the same random seed, and reused throughout the analysis so that the same instances were consistently evaluated across all models, explanation technique and metrics. Keeping the sampled instances fixed is important because it removes an additional source of variability and ensures a fair way of comparing the performance of explanations under the same conditions.

Once the evaluation subsets had been defined, the three explanation techniques were applied to each selected sample. This produced a local explanation for every instance, technique, and predictive model under analysis. The resulting attribution vectors were then passed to Quantus, which was used to compute the selected evaluation metrics. In this way, each sample generated a set of explanation scores for faithfulness, robustness, and complexity, allowing a direct comparison between techniques under the same conditions.

Finally, the metric values obtained for each sample were aggregated within each performance bin and consensus group. Minimum, mean, and maximum values were reported in order to capture not only the central tendency of explanation behavior, but also its possible best and worst-case behavior. This aggregation made it possible to identify whether explanation quality changes with higher or lower model performance, and whether those changes differ between consensus-correct and consensus-wrong samples.

\section{Results and Discussion} \label{results}

Due to the suitability of the F1-Score to balance both the precision and recall of a model, the obtained results were organized according to this model performance metric. Tables \ref{faithRes}--\ref{compRes} summarize the variation of explanation quality metrics across models with different ranges of F1-scores, considering 5\% bins, from 50\% to 100\%.

Within the tables, each cell reports the minimum, maximum and mean value for each explanation technique across models belonging to each bin and each consensus group. To ensure that a lower value is better in all five metrics, the Faithfulness metric was inverted. Additionally, color gradients were used to highlight the best results for each technique.

\begin{table*}[t]
\centering

\caption{Faithfulness results per model performance bin}
\label{faithRes}
\setlength{\tabcolsep}{3.5pt}
\scriptsize
\resizebox{\textwidth}{!}{%
\begin{tabular}{l c ccc ccc ccc ccc ccc ccc}
\toprule
\textbf{F1-Score}
& \textbf{Count}
& \multicolumn{9}{c}{\textbf{Consensus Correct}}
& \multicolumn{9}{c}{\textbf{Consensus Wrong}} \\
\cmidrule(lr){3-11} \cmidrule(lr){12-20}
&
& \multicolumn{3}{c}{Lime}
& \multicolumn{3}{c}{Kernel SHAP}
& \multicolumn{3}{c}{Feature Ablation}
& \multicolumn{3}{c}{Lime}
& \multicolumn{3}{c}{Kernel SHAP}
& \multicolumn{3}{c}{Feature Ablation} \\
\cmidrule(lr){3-5} \cmidrule(lr){6-8} \cmidrule(lr){9-11}
\cmidrule(lr){12-14} \cmidrule(lr){15-17} \cmidrule(lr){18-20}
&
&
min & mean & max &
min & mean & max &
min & mean & max &
min & mean & max &
min & mean & max &
min & mean & max \\
50-55 & 180 & \LIME{15}{-0.571} & \LIME{16}{-0.081} & \LIME{0}{\textbf{0.537}} & \KSHAP{21}{-0.765} & \KSHAP{60}{\textbf{-0.566}} & \KSHAP{31}{-0.233} & \FA{31}{-0.996} & \FA{19}{-0.882} & \FA{28}{-0.748} & \LIME{12}{-0.358} & \LIME{0}{\textbf{0.096}} & \LIME{2}{0.578} & \KSHAP{20}{-0.674} & \KSHAP{60}{\textbf{-0.529}} & \KSHAP{60}{\textbf{-0.419}} & \FA{34}{-0.999} & \FA{3}{-0.807} & \FA{6}{-0.506} \\
55-60 & 225 &  \LIME{11}{-0.488} & \LIME{23}{-0.162} & \LIME{19}{0.125} & \KSHAP{32}{-0.935} & \KSHAP{21}{-0.454} & \KSHAP{15}{-0.121} & \FA{0}{\textbf{-0.962}} & \FA{0}{\textbf{-0.809}} & \FA{15}{-0.619} & \LIME{2}{-0.134} & \LIME{7}{-0.017} & \LIME{19}{0.061} & \KSHAP{30}{-0.885} & \KSHAP{30}{-0.482} & \KSHAP{21}{-0.262} & \FA{0}{\textbf{-0.962}} & \FA{4}{-0.816} & \FA{16}{-0.662} \\
60-65 & 630 & \LIME{22}{-0.714} & \LIME{11}{-0.033} & \LIME{8}{0.373} & \KSHAP{31}{-0.917} & \KSHAP{33}{-0.552} & \KSHAP{15}{-0.120} & \FA{34}{-0.999} & \FA{27}{-0.914} & \FA{26}{-0.732} & \LIME{10}{-0.322} & \LIME{4}{0.028} & \LIME{0}{\textbf{0.635}} & \KSHAP{35}{-0.976} & \KSHAP{31}{-0.489} & \KSHAP{8}{-0.103} & \FA{33}{-0.998} & \FA{9}{-0.844} & \FA{0}{\textbf{-0.404}} \\
65-70 & 630 & \LIME{60}{\textbf{-0.979}} & \LIME{25}{-0.184} & \LIME{9}{0.336} & \KSHAP{32}{-0.931} & \KSHAP{20}{-0.439} & \KSHAP{9}{-0.081} & \FA{34}{-0.999} & \FA{10}{-0.848} & \FA{12}{-0.590} & \LIME{34}{-0.880} & \LIME{14}{-0.135} & \LIME{10}{0.330} & \KSHAP{33}{-0.942} & \KSHAP{27}{-0.458} & \KSHAP{6}{-0.089} & \FA{33}{-0.998} & \FA{10}{-0.849} & \FA{7}{-0.523} \\
70-75 & 720 & \LIME{34}{-0.956} & \LIME{27}{-0.196} & \LIME{12}{0.288} & \KSHAP{60}{\textbf{-0.980}} & \KSHAP{21}{-0.450} & \KSHAP{12}{-0.099} & \FA{60}{\textbf{-1.000}} & \FA{25}{-0.905} & \FA{3}{-0.506} & \LIME{60}{\textbf{-0.906}} & \LIME{14}{-0.138} & \LIME{15}{0.193} & \KSHAP{60}{\textbf{-0.984}} & \KSHAP{28}{-0.461} & \KSHAP{8}{-0.105} & \FA{60}{\textbf{-1.000}} & \FA{15}{-0.875} & \FA{5}{-0.484} \\
75-80 & 315 & \LIME{20}{-0.672} & \LIME{60}{\textbf{-0.284}} & \LIME{23}{0.057} & \KSHAP{30}{-0.896} & \KSHAP{8}{-0.344} & \KSHAP{0}{\textbf{-0.017}} & \FA{34}{-0.999} & \FA{60}{\textbf{-0.944}} & \FA{32}{-0.787} & \LIME{9}{-0.301} & \LIME{12}{-0.112} & \LIME{19}{0.062} & \KSHAP{27}{-0.824} & \KSHAP{12}{-0.315} & \KSHAP{0}{\textbf{-0.015}} & \FA{34}{-0.999} & \FA{27}{-0.941} & \FA{27}{-0.837} \\
80-85 & 255 & \LIME{13}{-0.534} & \LIME{19}{-0.116} & \LIME{4}{0.442} & \KSHAP{12}{-0.616} & \KSHAP{17}{-0.419} & \KSHAP{25}{-0.190} & \FA{60}{\textbf{-1.000}} & \FA{6}{-0.833} & \FA{18}{-0.652} & \LIME{15}{-0.421} & \LIME{6}{-0.007} & \LIME{5}{0.490} & \KSHAP{15}{-0.571} & \KSHAP{22}{-0.411} & \KSHAP{22}{-0.266} & \FA{60}{\textbf{-1.000}} & \FA{0}{\textbf{-0.792}} & \FA{9}{-0.541} \\
85-90 & 270 & \LIME{0}{\textbf{-0.260}} & \LIME{0}{\textbf{0.082}} & \LIME{8}{0.359} & \KSHAP{3}{-0.467} & \KSHAP{12}{-0.374} & \KSHAP{60}{\textbf{-0.263}} & \FA{26}{-0.990} & \FA{25}{-0.907} & \FA{60}{\textbf{-0.816}} & \LIME{0}{\textbf{-0.077}} & \LIME{4}{0.023} & \LIME{17}{0.120} & \KSHAP{17}{-0.615} & \KSHAP{17}{-0.368} & \KSHAP{17}{-0.208} & \FA{30}{-0.995} & \FA{28}{-0.946} & \FA{25}{-0.796} \\
90-95 & 360 & \LIME{2}{-0.306} & \LIME{20}{-0.129} & \LIME{25}{0.006} & \KSHAP{1}{-0.449} & \KSHAP{1}{-0.283} & \KSHAP{2}{-0.030} & \FA{60}{\textbf{-1.000}} & \FA{4}{-0.824} & \FA{0}{\textbf{-0.477}} & \LIME{5}{-0.188} & \LIME{11}{-0.098} & \LIME{17}{0.122} & \KSHAP{14}{-0.549} & \KSHAP{13}{-0.324} & \KSHAP{7}{-0.099} & \FA{60}{\textbf{-1.000}} & \FA{9}{-0.844} & \FA{2}{-0.431} \\
95-100 & 180 &  \LIME{1}{-0.284} & \LIME{32}{-0.254} & \LIME{60}{\textbf{-0.209}} & \KSHAP{0}{\textbf{-0.426}} & \KSHAP{0}{\textbf{-0.276}} & \KSHAP{24}{-0.187} & \FA{60}{\textbf{-1.000}} & \FA{26}{-0.909} & \FA{25}{-0.720} & \LIME{21}{-0.563} & \LIME{60}{\textbf{-0.502}} & \LIME{60}{\textbf{-0.424}} & \KSHAP{0}{\textbf{-0.262}} & \KSHAP{0}{\textbf{-0.209}} & \KSHAP{14}{-0.179} & \FA{60}{\textbf{-1.000}} & \FA{60}{\textbf{-0.987}} & \FA{60}{\textbf{-0.962}} \\
\bottomrule
\end{tabular}
}
\end{table*}

Table \ref{faithRes} reveals several recurring patterns in Faithfulness Estimate. First, Feature Ablation often reaches very high values, frequently in the [0.98, 1.00] range. This suggests that both the explanation technique and the evaluation metric rely on similar perturbation logic, which can significantly increase correlation.

Second, while Kernel SHAP and Feature Ablation are mostly positive, LIME appears more unstable, with the highest standard deviation among the three techniques (0.33). This behavior is likely related to LIME's sensitivity to locality hyper-parameters (e.g., kernel width): very narrow neighborhoods may not capture sufficient information about the model's behavior, whereas overly broad neighborhoods reduce locality and can degrade attribution quality.

Finally, consensus-wrong bins tend to produce lower faithfulness than consensus-correct bins, especially for LIME, suggesting that explanation reliability decreases in harder or less consistent prediction regimes.

\begin{table*}[t]
\centering
\caption{Selectivity results per model performance bin}
\label{selectRes}
\setlength{\tabcolsep}{4.5pt}
\scriptsize
\resizebox{\textwidth}{!}{%
\begin{tabular}{l c ccc ccc ccc ccc ccc ccc}
\toprule
\textbf{F1-Score}
& \textbf{Count}
& \multicolumn{9}{c}{\textbf{Consensus Correct}}
& \multicolumn{9}{c}{\textbf{Consensus Wrong}} \\
\cmidrule(lr){3-11} \cmidrule(lr){12-20}
&
& \multicolumn{3}{c}{Lime}
& \multicolumn{3}{c}{Kernel SHAP}
& \multicolumn{3}{c}{Feature Ablation}
& \multicolumn{3}{c}{Lime}
& \multicolumn{3}{c}{Kernel SHAP}
& \multicolumn{3}{c}{Feature Ablation} \\
\cmidrule(lr){3-5} \cmidrule(lr){6-8} \cmidrule(lr){9-11}
\cmidrule(lr){12-14} \cmidrule(lr){15-17} \cmidrule(lr){18-20}
&
&
min & mean & max &
min & mean & max &
min & mean & max &
min & mean & max &
min & mean & max &
min & mean & max \\

50-55 & 180 & \LIME{1}{0.527} & \LIME{0}{\textbf{0.632}} & \LIME{0}{0.724} & \KSHAP{0}{\textbf{0.548}} & \KSHAP{0}{\textbf{0.634}} & \KSHAP{0}{0.723} & \FA{0}{\textbf{0.544}} & \FA{0}{\textbf{0.635}} & \FA{0}{0.723} & \LIME{0}{\textbf{0.546}} & \LIME{0}{\textbf{0.634}} & \LIME{0}{0.723} & \KSHAP{0}{\textbf{0.541}} & \KSHAP{0}{\textbf{0.618}} & \KSHAP{0}{\textbf{0.721}} & \FA{0}{\textbf{0.559}} & \FA{0}{\textbf{0.637}} & \FA{0}{\textbf{0.724}} \\
55-60 & 225 & \LIME{0}{\textbf{0.537}} & \LIME{1}{0.619} & \LIME{1}{0.717} & \KSHAP{3}{0.519} & \KSHAP{2}{0.618} & \KSHAP{1}{0.719} & \FA{4}{0.509} & \FA{2}{0.611} & \FA{1}{0.716} & \LIME{5}{0.507} & \LIME{1}{0.625} & \LIME{1}{0.716} & \KSHAP{3}{0.518} & \KSHAP{0}{\textbf{0.618}} & \KSHAP{1}{0.705} & \FA{5}{0.521} & \FA{2}{0.619} & \FA{1}{0.715} \\
60-65 & 630 & \LIME{16}{0.399} & \LIME{11}{0.523} & \LIME{0}{\textbf{0.728}} & \KSHAP{18}{0.377} & \KSHAP{11}{0.527} & \KSHAP{0}{\textbf{0.725}} & \FA{17}{0.382} & \FA{12}{0.517} & \FA{0}{\textbf{0.725}} & \LIME{22}{0.388} & \LIME{10}{0.536} & \LIME{0}{\textbf{0.727}} & \KSHAP{24}{0.370} & \KSHAP{9}{0.537} & \KSHAP{1}{0.712} & \FA{26}{0.365} & \FA{11}{0.533} & \FA{0}{0.719} \\
65-70 & 630 & \LIME{21}{0.351} & \LIME{14}{0.503} & \LIME{15}{0.578} & \KSHAP{20}{0.354} & \KSHAP{12}{0.512} & \KSHAP{13}{0.589} & \FA{21}{0.337} & \FA{13}{0.502} & \FA{13}{0.583} & \LIME{26}{0.356} & \LIME{12}{0.521} & \LIME{6}{0.656} & \KSHAP{26}{0.358} & \KSHAP{11}{0.520} & \KSHAP{7}{0.644} & \FA{27}{0.353} & \FA{12}{0.521} & \FA{5}{0.669} \\
70-75 & 720 &  \LIME{23}{0.337} & \LIME{14}{0.495} & \LIME{13}{0.593} & \KSHAP{21}{0.347} & \KSHAP{14}{0.500} & \KSHAP{14}{0.585} & \FA{22}{0.333} & \FA{14}{0.488} & \FA{14}{0.568} & \LIME{29}{0.333} & \LIME{13}{0.513} & \LIME{8}{0.627} & \KSHAP{27}{0.346} & \KSHAP{12}{0.515} & \KSHAP{9}{0.621} & \FA{29}{0.338} & \FA{13}{0.510} & \FA{9}{0.612} \\
75-80 & 315 &\LIME{6}{0.481} & \LIME{13}{0.505} & \LIME{20}{0.527} & \KSHAP{4}{0.506} & \KSHAP{11}{0.528} & \KSHAP{16}{0.566} & \FA{6}{0.483} & \FA{13}{0.503} & \FA{18}{0.525} & \LIME{7}{0.495} & \LIME{12}{0.517} & \LIME{15}{0.539} & \KSHAP{6}{0.501} & \KSHAP{11}{0.523} & \KSHAP{15}{0.553} & \FA{9}{0.492} & \FA{14}{0.504} & \FA{17}{0.518} \\
80-85 & 255 & \LIME{19}{0.367} & \LIME{20}{0.446} & \LIME{16}{0.565} & \KSHAP{18}{0.376} & \KSHAP{19}{0.448} & \KSHAP{14}{0.585} & \FA{20}{0.354} & \FA{19}{0.440} & \FA{12}{0.594} & \LIME{32}{0.313} & \LIME{19}{0.446} & \LIME{8}{0.626} & \KSHAP{30}{0.328} & \KSHAP{19}{0.446} & \KSHAP{3}{0.682} & \FA{29}{0.337} & \FA{19}{0.448} & \FA{4}{0.680} \\
85-90 & 270 & \LIME{4}{0.504} & \LIME{9}{0.548} & \LIME{14}{0.587} & \KSHAP{4}{0.507} & \KSHAP{7}{0.563} & \KSHAP{12}{0.608} & \FA{4}{0.501} & \FA{9}{0.542} & \FA{13}{0.579} & \LIME{5}{0.513} & \LIME{8}{0.559} & \LIME{11}{0.597} & \KSHAP{2}{0.528} & \KSHAP{5}{0.570} & \KSHAP{10}{0.601} & \FA{5}{0.519} & \FA{8}{0.555} & \FA{11}{0.596} \\
90-95 & 360 & \LIME{17}{0.383} & \LIME{10}{0.538} & \LIME{7}{0.658} & \KSHAP{4}{0.507} & \KSHAP{6}{0.571} & \KSHAP{3}{0.694} & \FA{16}{0.385} & \FA{10}{0.530} & \FA{6}{0.660} & \LIME{9}{0.478} & \LIME{12}{0.515} & \LIME{14}{0.559} & \KSHAP{8}{0.486} & \KSHAP{10}{0.529} & \KSHAP{12}{0.578} & \FA{24}{0.380} & \FA{13}{0.514} & \FA{10}{0.604} \\
95-100 & 180 & \LIME{60}{\textbf{0.228}} & \LIME{60}{\textbf{0.299}} & \LIME{60}{\textbf{0.371}} & \KSHAP{60}{\textbf{0.210}} & \KSHAP{60}{\textbf{0.291}} & \KSHAP{60}{\textbf{0.371}} & \FA{60}{\textbf{0.206}} & \FA{60}{\textbf{0.280}} & \FA{60}{\textbf{0.341}} & \LIME{60}{\textbf{0.293}} & \LIME{60}{\textbf{0.296}} & \LIME{60}{\textbf{0.299}} & \KSHAP{60}{\textbf{0.291}} & \KSHAP{60}{\textbf{0.309}} & \KSHAP{60}{\textbf{0.318}} & \FA{60}{\textbf{0.293}} & \FA{60}{\textbf{0.297}} & \FA{60}{\textbf{0.300}} \\
\bottomrule
\end{tabular}
}
\end{table*}

Table \ref{selectRes} suggests that Selectivity is relatively consistent across explainability techniques for most F1 bins, with mean values usually concentrated between approximately 0.44 and 0.64. In the lower and mid-performance ranges (50--80), LIME, Kernel SHAP, and Feature Ablation show very similar behavior, indicating no clear dominant technique under this metric. A notable pattern appears in the highest bin (95--100), where Selectivity drops for all techniques (means around 0.28--0.31), suggesting that in these cases the ranked feature-removal process produces a smaller relative change in model output. Differences between Consensus Correct and Consensus Wrong are generally very small. Overall, Selectivity appears less discriminative between techniques than Faithfulness Estimate, but a useful faithfulness metric to rightfully compare Feature Ablation with the other two explanations.

\begin{table*}[t]
\centering
\caption{Average Sensitivity results per model performance bin}
\label{avgSensRes}
\setlength{\tabcolsep}{4.5pt}
\scriptsize
\resizebox{\textwidth}{!}{%
\begin{tabular}{l c ccc ccc ccc ccc ccc ccc}
\toprule
\textbf{F1-Score}
& \textbf{Count}
& \multicolumn{9}{c}{\textbf{Consensus Correct}}
& \multicolumn{9}{c}{\textbf{Consensus Wrong}} \\
\cmidrule(lr){3-11} \cmidrule(lr){12-20}
&
& \multicolumn{3}{c}{Lime}
& \multicolumn{3}{c}{Kernel SHAP}
& \multicolumn{3}{c}{Feature Ablation}
& \multicolumn{3}{c}{Lime}
& \multicolumn{3}{c}{Kernel SHAP}
& \multicolumn{3}{c}{Feature Ablation} \\
\cmidrule(lr){3-5} \cmidrule(lr){6-8} \cmidrule(lr){9-11}
\cmidrule(lr){12-14} \cmidrule(lr){15-17} \cmidrule(lr){18-20}
&
&
min & mean & max &
min & mean & max &
min & mean & max &
min & mean & max &
min & mean & max &
min & mean & max \\

50-55 & 180 & \LIME{34}{0.021} & \LIME{33}{0.134} & \LIME{27}{0.406} & \KSHAP{60}{\textbf{0.077}} & \KSHAP{30}{1.046} & \KSHAP{20}{1.572} & \FA{60}{\textbf{0.000}} & \FA{60}{\textbf{0.067}} & \FA{60}{\textbf{0.129}} & \LIME{60}{\textbf{0.011}} & \LIME{34}{0.134} & \LIME{27}{0.405} & \KSHAP{60}{\textbf{0.068}} & \KSHAP{60}{\textbf{0.889}} & \KSHAP{60}{\textbf{1.277}} & \FA{60}{\textbf{0.000}} & \FA{32}{0.057} & \FA{33}{0.124} \\
55-60 & 225 & \LIME{33}{0.027} & \LIME{34}{0.132} & \LIME{30}{0.314} & \KSHAP{30}{0.261} & \KSHAP{21}{1.151} & \KSHAP{12}{1.710} & \FA{11}{0.043} & \FA{13}{0.424} & \FA{25}{0.942} & \LIME{33}{0.026} & \LIME{33}{0.140} & \LIME{30}{0.324} & \KSHAP{33}{0.145} & \KSHAP{25}{1.048} & \KSHAP{30}{1.342} & \FA{0}{\textbf{0.058}} & \FA{12}{0.305} & \FA{16}{0.721} \\
60-65 & 630 & \LIME{35}{0.018} & \LIME{31}{0.160} & \LIME{17}{0.702} & \KSHAP{35}{0.092} & \KSHAP{60}{\textbf{0.985}} & \KSHAP{13}{1.693} & \FA{33}{0.003} & \FA{28}{0.184} & \FA{28}{0.732} & \LIME{34}{0.017} & \LIME{31}{0.159} & \LIME{17}{0.700} & \KSHAP{33}{0.148} & \KSHAP{33}{0.920} & \KSHAP{5}{1.708} & \FA{32}{0.005} & \FA{18}{0.220} & \FA{0}{\textbf{1.289}} \\
65-70 & 630 & \LIME{33}{0.027} & \LIME{27}{0.199} & \LIME{22}{0.558} & \KSHAP{29}{0.276} & \KSHAP{28}{1.071} & \KSHAP{25}{1.473} & \FA{32}{0.006} & \FA{25}{0.231} & \FA{27}{0.817} & \LIME{33}{0.026} & \LIME{27}{0.201} & \LIME{22}{0.566} & \KSHAP{31}{0.215} & \KSHAP{25}{1.051} & \KSHAP{24}{1.431} & \FA{33}{0.004} & \FA{15}{0.264} & \FA{9}{0.978} \\
70-75 & 720 & \LIME{60}{\textbf{0.016}} & \LIME{30}{0.170} & \LIME{24}{0.495} & \KSHAP{34}{0.118} & \KSHAP{24}{1.110} & \KSHAP{11}{1.739} & \FA{60}{\textbf{0.000}} & \FA{31}{0.134} & \FA{26}{0.912} & \LIME{34}{0.020} & \LIME{29}{0.178} & \LIME{24}{0.498} & \KSHAP{33}{0.155} & \KSHAP{21}{1.117} & \KSHAP{0}{\textbf{1.773}} & \FA{60}{\textbf{0.000}} & \FA{25}{0.140} & \FA{13}{0.829} \\
75-80 & 315 & \LIME{27}{0.062} & \LIME{0}{\textbf{0.454}} & \LIME{0}{\textbf{1.199}} & \KSHAP{25}{0.437} & \KSHAP{22}{1.142} & \KSHAP{24}{1.484} & \FA{60}{\textbf{0.000}} & \FA{27}{0.203} & \FA{31}{0.433} & \LIME{28}{0.061} & \LIME{0}{\textbf{0.454}} & \LIME{0}{\textbf{1.207}} & \KSHAP{23}{0.531} & \KSHAP{18}{1.161} & \KSHAP{17}{1.538} & \FA{60}{\textbf{0.000}} & \FA{20}{0.197} & \FA{24}{0.446} \\
80-85 & 255 & \LIME{26}{0.067} & \LIME{60}{\textbf{0.118}} & \LIME{60}{\textbf{0.179}} & \KSHAP{24}{0.478} & \KSHAP{19}{1.180} & \KSHAP{0}{\textbf{1.937}} & \FA{60}{\textbf{0.000}} & \FA{11}{0.460} & \FA{11}{2.134} & \LIME{25}{0.077} & \LIME{60}{\textbf{0.120}} & \LIME{60}{\textbf{0.177}} & \KSHAP{23}{0.511} & \KSHAP{22}{1.105} & \KSHAP{17}{1.534} & \FA{60}{\textbf{0.000}} & \FA{15}{0.268} & \FA{12}{0.862} \\
85-90 & 270 & \LIME{15}{0.129} & \LIME{22}{0.242} & \LIME{28}{0.372} & \KSHAP{3}{1.208} & \KSHAP{13}{1.251} & \KSHAP{60}{\textbf{1.287}} & \FA{0}{\textbf{0.064}} & \FA{31}{0.135} & \FA{34}{0.238} & \LIME{24}{0.082} & \LIME{25}{0.217} & \LIME{28}{0.374} & \KSHAP{9}{1.064} & \KSHAP{7}{1.350} & \KSHAP{9}{1.646} & \FA{15}{0.033} & \FA{24}{0.157} & \FA{28}{0.310} \\
90-95 & 360 & \LIME{26}{0.066} & \LIME{28}{0.181} & \LIME{30}{0.328} & \KSHAP{4}{1.171} & \KSHAP{3}{1.362} & \KSHAP{12}{1.718} & \FA{60}{\textbf{0.000}} & \FA{0}{\textbf{0.641}} & \FA{0}{\textbf{3.014}} & \LIME{27}{0.065} & \LIME{24}{0.221} & \LIME{20}{0.621} & \KSHAP{8}{1.096} & \KSHAP{7}{1.340} & \KSHAP{23}{1.442} & \FA{60}{\textbf{0.000}} & \FA{0}{\textbf{0.445}} & \FA{8}{0.991} \\
95-100 & 180 & \LIME{0}{\textbf{0.209}} & \LIME{8}{0.374} & \LIME{17}{0.716} & \KSHAP{0}{\textbf{1.303}} & \KSHAP{0}{\textbf{1.399}} & \KSHAP{21}{1.553} & \FA{60}{\textbf{0.000}} & \FA{34}{0.078} & \FA{35}{0.165} & \LIME{0}{\textbf{0.247}} & \LIME{2}{0.438} & \LIME{16}{0.727} & \KSHAP{0}{\textbf{1.410}} & \KSHAP{0}{\textbf{1.456}} & \KSHAP{16}{1.548} & \FA{60}{\textbf{0.000}} & \FA{60}{\textbf{0.019}} & \FA{60}{\textbf{0.056}} \\
\bottomrule
\end{tabular}
}
\end{table*}

Table \ref{avgSensRes} indicates that LIME is generally the most stable technique under small perturbations, with lower mean values across most F1 bins in both consensus groups. In contrast, Kernel SHAP presents consistently higher average sensitivity, especially in mid-to-high bins. For example, in the 90--95 bin under Consensus Correct, Kernel SHAP reaches a mean of 1.362, while LIME remains at 0.181.

Feature Ablation shows mixed behavior: it is often competitive in lower bins (e.g., 50--55, Consensus Correct, mean 0.067), but also shows sharp spikes in specific settings (e.g., 90--95, Consensus Correct, max 3.014). This suggests that FA can be robust on average but occasionally sensitive to particular feature perturbations.

\begin{table*}[t]
\centering
\caption{Max Sensitivity results per model performance bin}
\label{maxSensRes}
\setlength{\tabcolsep}{4.5pt}
\scriptsize
\resizebox{\textwidth}{!}{%
\begin{tabular}{l c ccc ccc ccc ccc ccc ccc}
\toprule
\textbf{F1-Score}
& \textbf{Count}
& \multicolumn{9}{c}{\textbf{Consensus Correct}}
& \multicolumn{9}{c}{\textbf{Consensus Wrong}} \\
\cmidrule(lr){3-11} \cmidrule(lr){12-20}
&
& \multicolumn{3}{c}{Lime}
& \multicolumn{3}{c}{Kernel SHAP}
& \multicolumn{3}{c}{Feature Ablation}
& \multicolumn{3}{c}{Lime}
& \multicolumn{3}{c}{Kernel SHAP}
& \multicolumn{3}{c}{Feature Ablation} \\
\cmidrule(lr){3-5} \cmidrule(lr){6-8} \cmidrule(lr){9-11}
\cmidrule(lr){12-14} \cmidrule(lr){15-17} \cmidrule(lr){18-20}
&
&
min & mean & max &
min & mean & max &
min & mean & max &
min & mean & max &
min & mean & max &
min & mean & max \\

50-55 & 180 & \LIME{30}{0.057} & \LIME{60}{\textbf{0.132}} & \LIME{33}{0.339} & \KSHAP{60}{\textbf{0.208}} & \KSHAP{24}{1.875} & \KSHAP{30}{2.965} & \FA{20}{0.186} & \FA{60}{\textbf{0.697}} & \FA{60}{\textbf{1.136}} & \LIME{35}{0.032} & \LIME{60}{\textbf{0.134}} & \LIME{60}{\textbf{0.340}} & \KSHAP{60}{\textbf{0.280}} & \KSHAP{31}{1.439} & \KSHAP{32}{2.312} & \FA{15}{0.245} & \FA{34}{0.727} & \FA{35}{1.165} \\
55-60 & 225 &\LIME{32}{0.042} & \LIME{35}{0.138} & \LIME{60}{\textbf{0.287}} & \KSHAP{29}{0.406} & \KSHAP{0}{\textbf{3.079}} & \KSHAP{10}{7.977} & \FA{16}{0.231} & \FA{21}{2.848} & \FA{26}{7.804} & \LIME{33}{0.044} & \LIME{30}{0.186} & \LIME{33}{0.464} & \KSHAP{33}{0.370} & \KSHAP{0}{\textbf{3.139}} & \KSHAP{0}{\textbf{8.261}} & \FA{8}{0.336} & \FA{27}{3.265} & \FA{29}{12.404} \\
60-65 & 630 & \LIME{34}{0.033} & \LIME{25}{0.270} & \LIME{1}{0.997} & \KSHAP{31}{0.331} & \KSHAP{60}{\textbf{1.342}} & \KSHAP{31}{2.843} & \FA{17}{0.214} & \FA{34}{0.810} & \FA{34}{1.726} & \LIME{34}{0.033} & \LIME{23}{0.263} & \LIME{22}{0.996} & \KSHAP{31}{0.451} & \KSHAP{60}{\textbf{1.206}} & \KSHAP{31}{2.477} & \FA{14}{0.261} & \FA{34}{0.760} & \FA{34}{2.319} \\
65-70 & 630 & \LIME{34}{0.031} & \LIME{21}{0.335} & \LIME{1}{0.996} & \KSHAP{26}{0.524} & \KSHAP{12}{2.466} & \KSHAP{0}{\textbf{10.575}} & \FA{12}{0.279} & \FA{26}{1.972} & \FA{16}{15.951} & \LIME{35}{0.032} & \LIME{16}{0.337} & \LIME{22}{0.997} & \KSHAP{31}{0.450} & \KSHAP{21}{1.999} & \KSHAP{14}{5.678} & \FA{0}{\textbf{0.431}} & \FA{34}{0.950} & \FA{34}{3.687} \\
70-75 & 720 &  \LIME{60}{\textbf{0.026}} & \LIME{18}{0.375} & \LIME{1}{0.997} & \KSHAP{34}{0.234} & \KSHAP{26}{1.765} & \KSHAP{24}{4.510} & \FA{22}{0.163} & \FA{34}{0.837} & \FA{34}{2.009} & \LIME{60}{\textbf{0.029}} & \LIME{12}{0.377} & \LIME{22}{0.996} & \KSHAP{33}{0.372} & \KSHAP{23}{1.844} & \KSHAP{18}{4.911} & \FA{15}{0.249} & \FA{34}{0.823} & \FA{35}{1.480} \\
75-80 & 315 & \LIME{27}{0.071} & \LIME{11}{0.466} & \LIME{1}{0.985} & \KSHAP{16}{0.875} & \KSHAP{35}{1.348} & \KSHAP{60}{\textbf{1.798}} & \FA{31}{0.053} & \FA{34}{0.798} & \FA{35}{1.291} & \LIME{28}{0.078} & \LIME{4}{0.466} & \LIME{22}{0.985} & \KSHAP{22}{0.766} & \KSHAP{33}{1.324} & \KSHAP{60}{\textbf{1.771}} & \FA{33}{0.027} & \FA{34}{0.814} & \FA{35}{1.214} \\
80-85 & 255 & \LIME{24}{0.094} & \LIME{23}{0.305} & \LIME{1}{0.990} & \KSHAP{19}{0.751} & \KSHAP{5}{2.818} & \KSHAP{19}{5.737} & \FA{60}{\textbf{0.000}} & \FA{0}{\textbf{5.892}} & \FA{0}{\textbf{28.289}} & \LIME{23}{0.117} & \LIME{19}{0.308} & \LIME{22}{0.994} & \KSHAP{17}{0.966} & \KSHAP{3}{2.989} & \KSHAP{1}{8.149} & \FA{60}{\textbf{0.000}} & \FA{0}{\textbf{12.798}} & \FA{0}{\textbf{72.610}} \\
85-90 & 270 & \LIME{0}{\textbf{0.233}} & \LIME{0}{\textbf{0.626}} & \LIME{0}{\textbf{1.016}} & \KSHAP{4}{1.267} & \KSHAP{32}{1.503} & \KSHAP{35}{1.812} & \FA{0}{\textbf{0.424}} & \FA{33}{1.009} & \FA{34}{1.676} & \LIME{26}{0.090} & \LIME{0}{\textbf{0.508}} & \LIME{0}{\textbf{2.086}} & \KSHAP{15}{1.041} & \KSHAP{28}{1.610} & \KSHAP{33}{2.147} & \FA{1}{0.420} & \FA{35}{0.611} & \FA{60}{\textbf{0.750}} \\
90-95 & 360 & \LIME{26}{0.080} & \LIME{32}{0.172} & \LIME{34}{0.315} & \KSHAP{5}{1.253} & \KSHAP{10}{2.592} & \KSHAP{16}{6.618} & \FA{60}{\textbf{0.000}} & \FA{16}{3.531} & \FA{12}{18.620} & \LIME{28}{0.077} & \LIME{22}{0.276} & \LIME{22}{1.013} & \KSHAP{15}{1.066} & \KSHAP{28}{1.615} & \KSHAP{34}{1.942} & \FA{60}{\textbf{0.000}} & \FA{33}{1.147} & \FA{34}{2.914} \\
95-100 & 180 & \LIME{3}{0.214} & \LIME{20}{0.344} & \LIME{23}{0.540} & \KSHAP{0}{\textbf{1.417}} & \KSHAP{25}{1.834} & \KSHAP{31}{2.677} & \FA{60}{\textbf{0.000}} & \FA{30}{1.483} & \FA{30}{4.686} & \LIME{0}{\textbf{0.278}} & \LIME{14}{0.360} & \LIME{32}{0.491} & \KSHAP{0}{\textbf{1.626}} & \KSHAP{22}{1.925} & \KSHAP{33}{2.171} & \FA{60}{\textbf{0.000}} & \FA{60}{\textbf{0.521}} & \FA{35}{1.053} \\
\bottomrule
\end{tabular}
}
\end{table*}

The Max-sensitivity metric in Table \ref{maxSensRes} highlights the worst-case behavior. LIME remains comparatively controlled across bins, while Kernel SHAP and Feature Ablation show higher peaks, especially in more challenging regions. Some outliers appear for Feature Ablation in the 80--85 bin (Consensus Wrong), where the maximum reaches 72.610, and for Kernel SHAP in the same region with means above 2.8. Additionally, while there is little correlation between the two consensus groups, it can be observed that the larger peaks referenced above are not as high in the Consensus Correct category.

These extreme values suggest that some explanation instances are highly sensitive to perturbations even when average behavior appears acceptable. This shows the importance of analyzing both average and maximum sensitivity: average metrics capture typical behavior, while max metrics expose rare but potentially critical instabilities.

\begin{table*}[t]
\centering
\caption{Complexity results per model performance bin}
\label{compRes}
\setlength{\tabcolsep}{4.5pt}
\scriptsize
\resizebox{\textwidth}{!}{%
\begin{tabular}{l c ccc ccc ccc ccc ccc ccc}
\toprule
\textbf{F1-Score}
& \textbf{Count}
& \multicolumn{9}{c}{\textbf{Consensus Correct}}
& \multicolumn{9}{c}{\textbf{Consensus Wrong}} \\
\cmidrule(lr){3-11} \cmidrule(lr){12-20}
&
& \multicolumn{3}{c}{Lime}
& \multicolumn{3}{c}{Kernel SHAP}
& \multicolumn{3}{c}{Feature Ablation}
& \multicolumn{3}{c}{Lime}
& \multicolumn{3}{c}{Kernel SHAP}
& \multicolumn{3}{c}{Feature Ablation} \\
\cmidrule(lr){3-5} \cmidrule(lr){6-8} \cmidrule(lr){9-11}
\cmidrule(lr){12-14} \cmidrule(lr){15-17} \cmidrule(lr){18-20}
&
&
min & mean & max &
min & mean & max &
min & mean & max &
min & mean & max &
min & mean & max &
min & mean & max \\

50-55 & 180 & \LIME{26}{1.321} & \LIME{33}{2.561} & \LIME{27}{4.316} & \KSHAP{34}{1.097} & \KSHAP{33}{2.702} & \KSHAP{27}{4.548} & \FA{19}{1.053} & \FA{27}{1.848} & \FA{60}{\textbf{2.455}} & \LIME{29}{1.320} & \LIME{33}{2.553} & \LIME{27}{4.320} & \KSHAP{33}{1.198} & \KSHAP{34}{2.635} & \KSHAP{30}{4.271} & \FA{19}{1.199} & \FA{29}{1.990} & \FA{32}{2.765} \\
55-60 & 225 & \LIME{17}{1.980} & \LIME{30}{2.705} & \LIME{60}{\textbf{3.537}} & \KSHAP{29}{1.487} & \KSHAP{32}{2.781} & \KSHAP{60}{\textbf{3.763}} & \FA{14}{1.234} & \FA{24}{2.006} & \FA{30}{3.036} & \LIME{24}{1.977} & \LIME{32}{2.682} & \LIME{60}{\textbf{3.535}} & \KSHAP{31}{1.522} & \KSHAP{33}{2.767} & \KSHAP{60}{\textbf{3.796}} & \FA{14}{1.371} & \FA{27}{2.101} & \FA{30}{3.040} \\
60-65 & 630 & \LIME{26}{1.296} & \LIME{28}{2.820} & \LIME{12}{5.914} & \KSHAP{60}{\textbf{1.038}} & \KSHAP{30}{2.900} & \KSHAP{11}{6.098} & \FA{22}{0.958} & \FA{31}{1.689} & \FA{35}{2.468} & \LIME{29}{1.310} & \LIME{30}{2.821} & \LIME{12}{5.932} & \KSHAP{60}{\textbf{1.021}} & \KSHAP{31}{2.901} & \KSHAP{11}{6.100} & \FA{24}{0.987} & \FA{34}{1.789} & \FA{60}{\textbf{2.450}} \\
65-70 & 630 & \LIME{60}{\textbf{0.635}} & \LIME{27}{2.896} & \LIME{23}{4.785} & \KSHAP{28}{1.582} & \KSHAP{25}{3.201} & \KSHAP{22}{5.037} & \FA{32}{0.621} & \FA{29}{1.769} & \FA{28}{3.300} & \LIME{60}{\textbf{0.628}} & \LIME{29}{2.898} & \LIME{24}{4.681} & \KSHAP{30}{1.636} & \KSHAP{27}{3.218} & \KSHAP{23}{4.936} & \FA{27}{0.870} & \FA{31}{1.944} & \FA{27}{3.383} \\
70-75 & 720 & \LIME{31}{0.904} & \LIME{18}{3.405} & \LIME{14}{5.733} & \KSHAP{31}{1.323} & \KSHAP{18}{3.613} & \KSHAP{11}{6.141} & \FA{60}{\textbf{0.524}} & \FA{23}{2.040} & \FA{28}{3.224} & \LIME{33}{0.917} & \LIME{22}{3.407} & \LIME{14}{5.745} & \KSHAP{34}{1.106} & \KSHAP{22}{3.560} & \KSHAP{11}{6.138} & \FA{60}{\textbf{0.555}} & \FA{27}{2.100} & \FA{29}{3.188} \\
75-80 & 315 &  \LIME{25}{1.368} & \LIME{0}{\textbf{4.427}} & \LIME{0}{\textbf{7.163}} & \KSHAP{29}{1.489} & \KSHAP{0}{\textbf{4.685}} & \KSHAP{0}{\textbf{7.191}} & \FA{17}{1.117} & \FA{0}{\textbf{3.103}} & \FA{0}{\textbf{6.531}} & \LIME{29}{1.333} & \LIME{8}{4.418} & \LIME{0}{\textbf{7.162}} & \KSHAP{29}{1.696} & \KSHAP{7}{4.707} & \KSHAP{0}{\textbf{7.150}} & \FA{19}{1.191} & \FA{0}{\textbf{3.182}} & \FA{0}{\textbf{6.501}} \\
80-85 & 255 & \LIME{28}{1.198} & \LIME{60}{\textbf{2.430}} & \LIME{34}{3.634} & \KSHAP{32}{1.295} & \KSHAP{60}{\textbf{2.570}} & \KSHAP{35}{3.796} & \FA{17}{1.115} & \FA{60}{\textbf{1.482}} & \FA{35}{2.481} & \LIME{30}{1.219} & \LIME{60}{\textbf{2.433}} & \LIME{34}{3.649} & \KSHAP{32}{1.316} & \KSHAP{60}{\textbf{2.589}} & \KSHAP{34}{3.861} & \FA{19}{1.203} & \FA{60}{\textbf{1.764}} & \FA{30}{3.076} \\
85-90 & 270 & \LIME{0}{\textbf{3.274}} & \LIME{15}{3.559} & \LIME{26}{4.484} & \KSHAP{2}{3.537} & \KSHAP{15}{3.780} & \KSHAP{26}{4.640} & \FA{0}{\textbf{1.691}} & \FA{18}{2.262} & \FA{30}{3.022} & \LIME{12}{3.308} & \LIME{20}{3.552} & \LIME{27}{4.394} & \KSHAP{14}{3.533} & \KSHAP{18}{3.832} & \KSHAP{26}{4.695} & \FA{0}{\textbf{1.934}} & \FA{14}{2.614} & \FA{25}{3.579} \\
90-95 & 360 & \LIME{7}{2.710} & \LIME{15}{3.572} & \LIME{25}{4.602} & \KSHAP{10}{2.948} & \KSHAP{14}{3.815} & \KSHAP{24}{4.802} & \FA{23}{0.933} & \FA{22}{2.066} & \FA{28}{3.310} & \LIME{17}{2.729} & \LIME{20}{3.537} & \LIME{26}{4.461} & \KSHAP{18}{2.971} & \KSHAP{19}{3.820} & \KSHAP{25}{4.766} & \FA{7}{1.647} & \FA{15}{2.591} & \FA{21}{4.045} \\
95-100 & 180 & \LIME{0}{3.263} & \LIME{4}{4.208} & \LIME{20}{5.078} & \KSHAP{0}{\textbf{3.664}} & \KSHAP{3}{4.486} & \KSHAP{20}{5.249} & \FA{18}{1.077} & \FA{18}{2.284} & \FA{30}{3.039} & \LIME{0}{\textbf{4.787}} & \LIME{0}{\textbf{4.970}} & \LIME{20}{5.084} & \KSHAP{0}{\textbf{5.143}} & \KSHAP{0}{\textbf{5.218}} & \KSHAP{19}{5.297} & \FA{3}{1.825} & \FA{15}{2.570} & \FA{29}{3.131} \\
\bottomrule
\end{tabular}
}
\end{table*}

Table \ref{compRes} shows a clear increase in complexity with model-performance bins for all techniques, particularly from 75--80 onward. This pattern suggests that higher-performing settings tend to distribute attribution over a larger set of features instead of relying on a small subset. However, this trend may be influenced by dataset characteristics. Datasets with a higher feature count achieved an overall higher score than those with a smaller amount, indicating that the number of features in the dataset could influence the reliability of explanations. This could indicate a possible reason as to why previous metrics didn't behave as expected. 

Across techniques, LIME and Kernel SHAP usually present higher complexity than Feature Ablation, especially in high bins. For instance, in the 95--100 bin (Consensus Wrong), mean complexity is 4.970 for LIME and 5.218 for Kernel SHAP, compared to 2.570 for Feature Ablation. This indicates that FA explanations are generally sparser, whereas LIME and Kernel SHAP produce more distributed attribution.

After the previous results were analyzed, it became important to further analyze the relationship between the number of features in the datasets and the value of each metric.

\begin{figure}[h]
    \centering
    \includegraphics[width=1\linewidth]{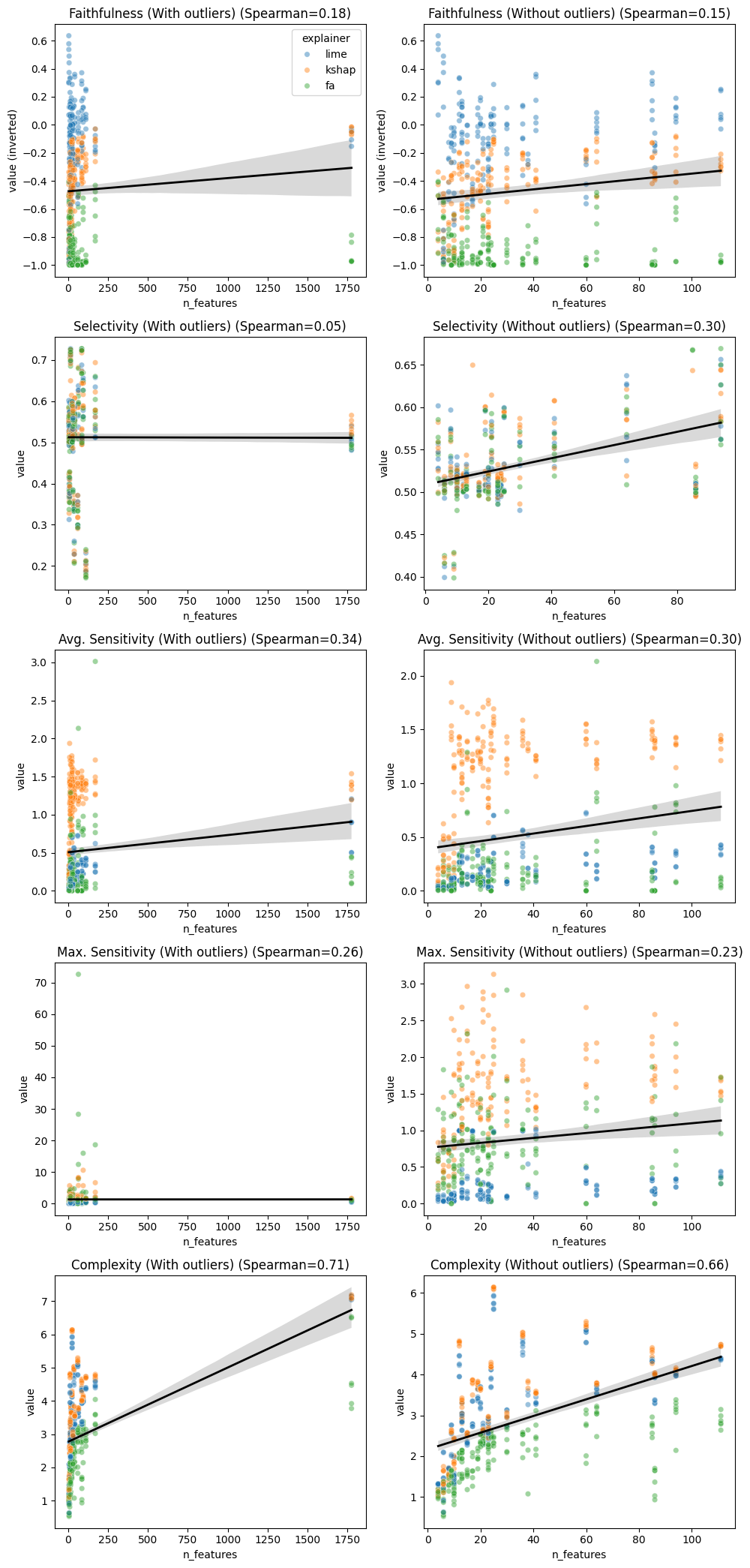}
    \caption{Correlation of explainability metrics with number of features of a tabular dataset.}
    \label{fig:avg_sens_corr}
\end{figure}

Fig. \ref{fig:avg_sens_corr} shows the correlation between the average sensitivity and faithfulness metrics and the number of features in each dataset. The left panels includes all datasets, whereas the right panel omits the only dataset with more than 150 features (1776 features) to improve readability. For these metrics, the relationship appears to be weak to moderate, suggesting that feature dimensionality influences the metric values but does not fully determine them. A similar trend is visible for the remaining metrics, with varying magnitudes, all showing positive correlations. Complexity presents the strongest association, with a correlation of 0.71. 

Overall, these results suggest that explanation quality is not primarily determined by model predictive performance alone. Instead, the behavior of the evaluated techniques appears to be more strongly influenced by the complexity of a model's decision boundaries, which in turn is influenced by the complexity of the dataset. Since this complexity is itself related to the structure and difficulty of the training data, especially in datasets with many features or more intricate patterns, the trustworthiness of local explanation techniques may be significantly affected by dataset complexity. In this sense, complex tabular datasets may pose an additional challenge for explanation reliability, even when the predictive performance of the underlying models is high.

\section{Conclusions} \label{conclusions}

This work investigated the trustworthiness of local explainability techniques when applied to complex tabular classification tasks, with the goal of identifying patterns and correlations between dataset complexity, model performance, and explanation reliability. In particular, the applicability of multiple robustness, faithfulness, and complexity metrics to tabular datasets was analyzed. A benchmark was carried out for three widely used local explainability techniques, LIME, Kernel SHAP, and Feature Ablation. The selected evaluation metrics were implemented using the Quantus library and the explanations were generated using Captum.

Results show a weak correlation of the F1-Score with the metrics value, sometimes showing an inverse trend to expectations. For example, the higher the F1-Score, the higher metrics like Average and Max sensitivity tend do be. A possible reason is that datasets with higher F1-Score being more complex as pointed out by the complexity metric, which suggests that variables like the number of features in the dataset could have relevant significance in determining explanation reliability. Furthermore, Consensus Correct samples seem to have a slightly higher value than Consensus Wrong samples, however there does not seem to be a direct correlation between the models correct classification and the reliability of explanations. In general, Lime seems to not be a reliable explanation with the tested kernel parameters, as the values can vary a lot and have low faithfulness. However, it is the most robust of the three explanations tested.
On the other hand, Kernel SHAP is the least robust of the explanations but in terms of worst-case scenarios it spikes less than Feature Ablation.

As a next step, it is important to investigate in more detail the correlation between dataset complexity, more specifically the number of features, and the values of the metrics. The goal being to evaluate if explanations deteriorate when features considered as "unimportant" by the explanation are removed one by one and models are trained using those reduced versions. This will enable a more detailed analysis into how consistent and faithful are the explanations to the decision boundaries of machine learning and deep learning models.

\section*{\uppercase{Acknowledgements}}
This work was supported and received funding from “CyberPRAISE - Cybersecurity research for PRivAte, Intelligent and truStablE solutions” - NORTE2030-FEDER-01820300.

\bibliographystyle{apalike}
{\small
\bibliography{references}}

\end{document}